\pdfoutput=1
\documentclass[11pt]{article}
\usepackage{xcolor}
\usepackage{colortbl}
\usepackage[]{acl}
\usepackage{times}
\usepackage{latexsym}
\usepackage{multirow}
\usepackage[T1]{fontenc}
\usepackage[utf8]{inputenc}

\usepackage{microtype}
\usepackage{graphicx}

\usepackage{tikz}
\newcommand\diag[4]{%
  \multicolumn{1}{|p{#2}|}{\hskip-\tabcolsep
  $\vcenter{\begin{tikzpicture}[baseline=0,anchor=south west,inner sep=#1]
  \path[use as bounding box] (0,0) rectangle (#2+2\tabcolsep,\baselineskip);
  \node[minimum width={#2+2\tabcolsep},minimum height=\baselineskip+\extrarowheight] (box) {};
  \draw (box.north west) -- (box.south east);
  \node[anchor=south west] at (box.south west) {#3};
  \node[anchor=north east] at (box.north east) {#4};
 \end{tikzpicture}}$\hskip-\tabcolsep}}

\title{Using BERT Embeddings to Model Word Importance in Conversational Transcripts for Deaf and Hard of Hearing Users}

\author{Akhter Al Amin, Saad Hassan, Cecilia O. Alm, Matt Huenerfauth \\
        Rochester Institute of Technology \\ 1 Lomb Memorial Drive, Rochester, NY \\ \texttt{\{aa7510,sh2513,coagla,matt.huenerfauth\}@rit.edu} }

\begin{document}
\maketitle
\begin{abstract}

Deaf and hard of hearing individuals regularly rely on captioning while watching live TV. Live TV captioning is evaluated by regulatory agencies using various caption evaluation metrics. However,  caption evaluation metrics are often not informed by preferences of DHH users or how meaningful the captions are.  There is a need to construct caption evaluation metrics that take the relative importance of words in a transcript into account. We conducted correlation analysis between two types of word embeddings and human-annotated labeled word-importance scores in existing corpus. We found that normalized contextualized word embeddings generated using BERT correlated better with manually annotated importance scores than word2vec-based word embeddings. We make available a pairing of word embeddings and their human-annotated importance scores. We also provide proof-of-concept utility by training word importance models, achieving an F1-score of 0.57 in the 6-class word importance classification task. 
\end{abstract}

\section{Introduction}

Over 360 million people worldwide are Deaf or Hard of Hearing (DHH) \cite{mitchell2006many, blanchfield2001severely}. In the U.S. alone, over 15\% people are DHH, and regularly rely on captioning while watching videos to perceive salient auditory information \cite{larwan2019}. To provide quality captioning services to this group, it is essential to monitor the quality of captioning regularly. Regulators, e.g., the Federal Communication Commission (FCC) in the U.S. \cite{fcc2014} are entrusted with regularly checking the quality of caption transcription generated by different broadcasters. However, given the abundant production of captioned live TV broadcasts, caption evaluation is a tedious and costly task.

DHH viewers are often dissatisfied with the quality of captioning provided in live contexts, which provide less time for caption production than pre-recorded contexts \cite{metric2021akhter, Kushalnagar2018}. If regulatory organizations that measure the quality of captions used quality metrics that better reflect the  DHH users' preferences, DHH viewers' experience may improve. 

%  (as could that of other viewers who take advantage of captions)

Existing metrics used in transcription or captioning include Word Error Rate (WER) \cite{WER_2018} or Number of Error in Recognition (NER) \cite{NER_2015}. As noted by \citet{highlight_kafle_2019}, a major shortcoming of these metrics is that they do not consider the importance of individual words when measuring the accuracy of captioned transcripts (comparing to the reference transcript) and most metrics assign equal weights to each word. DHH viewers rely more heavily on important keywords while skimming through caption text \cite{highlight_kafle_2019}. 

Motivated by these shortcomings, prior work had proposed metrics which assign differential importance weights to individual words in captioned text when calculating an evaluation score \cite{Kafle_ACE, Kafle2019}. Specifically, this prior work leveraged word2vec-based word embeddings to generate and propagate features to another layer of the network \cite{kafle-huenerfauth-2018-corpus}. We build on this prior work and propose an updated approach. The feature space we are using contains both contextual and semantic information of the captioned text, which is crucial in conversational setting, often common in TV, and may better capture long-distance semantic and syntactic relationships. Thus, in this work, we contribute more current strategies for calculating importance of words in transcript text, toward a metric that takes word-importance into account when evaluating captions. Our contributions in this paper include:
 
 \begin{enumerate}
     \item \textbf{We conducted a comparative correlation analysis between human-annotated importance scores for words in  conversational transcripts and aggregated lexical semantic score generated from: (a)  word2vec-based word embeddings as in prior work contrasted with (b) BERT-based contextualized embeddings}. Our findings revealed that scores generated from contextualized embeddings had higher correlation with the human-annotated word-importance scores.    
     \item  \textbf{We contribute data consisting of BERT contextualized word embeddings, paired with their word-importance scores, to augment a prior dataset of human-assigned importance scores for words in conversational transcripts \cite{kafle-huenerfauth-2018-corpus}.} This enhanced data can be used by researchers for constructing improved caption-evaluation metrics or by researchers studying conversational discourse.
     \item  \textbf{To illustrate the use of this dataset, we show how interpretable classical machine-learning models can be trained to determine the importance of words using these contextualized word embedding vectors from our data.} In this proof-of-concept study, we show how these data can be used in training models. We leave detailed evaluation and comparison of models for future work.
 \end{enumerate}

\section{Related Work}
\subsection{Word Importance Prediction}
NLP researchers have explored approaches to determine word-importance for various downstream tasks, e.g. term weight determination when querying text \cite{dai2020importance}, for text summarization  \cite{hong-nenkova-2014-improving} or text classification \cite{sheikh:hal-01331720}. Prior research on identifying and scoring important words in a text has largely focused on the task of keyword or important-term extraction \cite{dai2020importance,sheikh:hal-01331720}. This task involves identifying words in a document that densely summarize it. Several automatic keyword-extraction techniques have been investigated, including unsupervised methods such as interpolation of Term Frequency and Inverse Document Frequency (TF-IDF) weighting \cite{tf_idf_2010}, Positive Pointwise Mutual Information (PPMI) \cite{PPMI_Bouma}, word2vec embedding \cite{sheikh:hal-01331720}, and supervised methods that leverage linguistic features from text for word importance estimation \cite{dai2020importance, kafle-huenerfauth-2018-corpus}. While the conceptualization of word importance as a keyword-extraction problem has enabled retrieving relevant information from large textual or multimedia datasets \cite{dai2020importance, Shah02astudy}, this approach may not generalize across domains and functional, situational contexts of language use. For instance, given the meandering nature of topic transitions in television news broadcasts or talk shows \cite{Kafle_ACE}, when processing caption transcripts, a model of word importance that is more local may be more successful, rather than considering the entire transcript of the broadcast or show.  

\subsection{Caption Evaluation Methods}
Several caption evaluation approaches have been proposed \cite{WER_2018, NCAM}, with some approaches specifically taking into account the perspective of DHH participants \cite{kafle-huenerfauth-2018-corpus, metric2021akhter}. The most common caption evaluation used by different regulatory organizations is Word Error Rate (WER) \cite{WER_2018}. While penalizing insertion, deletion, and substitution errors in transcripts, a limitation of WER is that it considers importance of each word token equally. To address this, \citet{NCAM} proposed a metric that assign weights to words in a text, but this probabilistic approach has not been trained on weights set to address priorities assigned by actual caption users.  

In the most closely related work, \citet{kafle-huenerfauth-2018-corpus} investigated models for predicting word-importance during captioned one-on-one conversations. Their Automatic Caption Evaluation (ACE) framework utilized a variety of linguistic features to predict which words in a caption text were most important to its meaning, and which would be most problematic if incorrectly transcribed in a caption. Prior research on determining the importance of a word in a document had shown that an embedding can characterize a word’s syntactic (e.g., word dependencies) and semantic character (e.g., named entity labeling), which in turn can help estimate a word’s importance \cite{sheikh:hal-01331720}. Thus, \citet{kafle-huenerfauth-2018-corpus} used word2vec embeddings of words in the transcript. In this paper, we examine whether an alternative embedding, based on BERT, would lead to superior models of word-importance.

\subsection{Annotation of Word Importance Scores}
In this work, we contribute a dataset that augments a previously-released dataset from \citet{kafle-huenerfauth-2018-corpus}, consisting of a 25,000-token subset of the Switchboard corpus of conversational transcripts \cite{SWITCHBOARD_1992}.  \citet{kafle-huenerfauth-2018-corpus} asked a pair of human annotators to assign word-importance scores to each word within these transcripts, on a range from 0.0 to 1.0, where 1.0 was most important. After partitioning scores into 6 discrete categories: [0-0.1), [0.1-0.3), [0.3-0.5), [0.5-0.7), [0.7-0.9), and [0.9 - 1], they trained a Neural Network-based classifier, using Long Short Term Memory (LSTM), to predict the importance category of each word in these transcripts. We augment this annotated corpus with recent contextualized word embeddings from BERT \cite{devlin-etal-2019-bert}, pairing up the embeddings with the  hand-annotated word importance data.
\section{Corpus Augmentation}

\subsection{Extracting Word Embeddings Vectors}
We have augmented the dataset described above, and will be releasing the version that  includes two embeddings per word token: BERT contextualized word embeddings and word2vec embeddings. With this paper,
we will be releasing the BERT-generated contextualized word embeddings\footnote[1]{\url{https://nyu.databrary.org/volume/1447}} of 25,000 tokens, each with a feature vector of length 768, augmented with the human-annotated word-importance scores\footnote[2]{\url{http://latlab.ist.rit.edu/lrec2018/}}.

To enable comparison with the work of \citet{kafle-huenerfauth-2018-corpus}, we extracted a word2vec \cite{rehurek2011gensim} embedding vector of length 100 for each word that occurred at least twice within each transcript. Next, we employed the pre-trained BERT model entitled \emph{bert-base-uncased} \cite{devlin-etal-2019-bert} to generate a contextualized word-embedding vector for each word within transcripts. For each word within each sentence, using BERT, we generated a three-dimensional embedding of shape $32 \times 12 \times 768$. These embeddings were created based upon the architecture of the pre-trained BERT model that included 32 transformer blocks, 12 attention heads and 768 hidden layers. We follow prior work that has reshaped or composed the three dimensions into a one-dimensional vector while retaining similar semantic information \cite{turton2020deriving}. After performing these operations, for each word we obtained a contextualized embedding vector of length $768$.

\begin{figure}
\small
    \centering
    \begin{minipage}{0.23\textwidth}
        \includegraphics[width=0.95\textwidth]{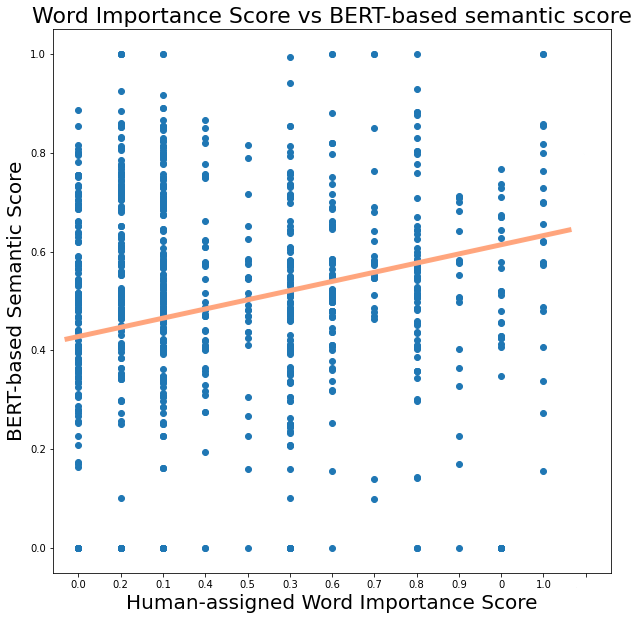}
    \caption*{(a)}
    \end{minipage}
    \begin{minipage}{0.23\textwidth}
        \includegraphics[width=0.97\textwidth, height=3.4cm]{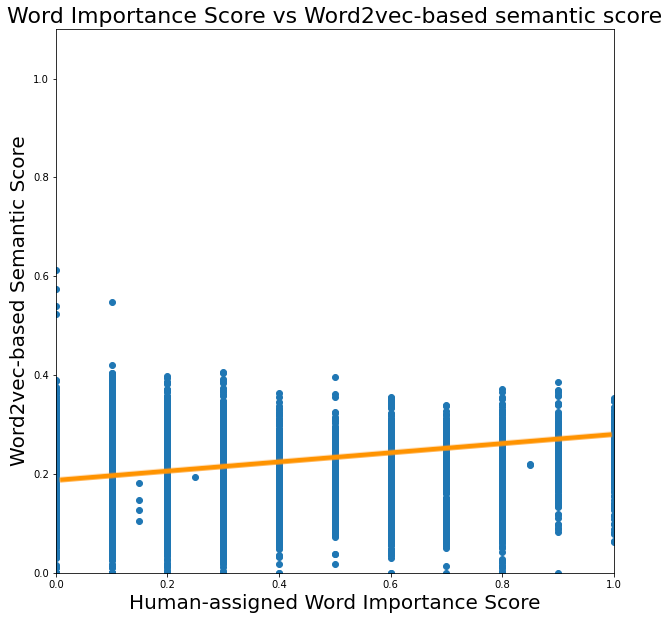}
    \caption*{(b)}
    \end{minipage}
    \caption{Scatter plots for (a) the human-annotated score vs. BERT embedding-based semantic score, and (b) the human-annotated score vs. the word2vec embedding-based semantic score. The first 1200 words from the dataset are shown. }
    \label{fig:correlation}
\end{figure}

\begin{table}[h]
\small
\centering

\begin{tabular}{|l|l|l|l|} 
\hline
    \diag{.1em}{2.1cm}{Method}{Word} &  \emph{sunday}  & \emph{noise} & \emph{plan}  \\
\hline
    Human-assigned score & 0.60  & 0.40 & 0.70 \\
\hline
    BERT & 0.10 & 0.42 & 0.61  \\
\hline

    word2vec & 0.35 & 0.17 & 0.18 \\
\hline

\end{tabular}
\caption{Three sample words, \emph{sunday}, \emph{noise}, and \emph{plan} have been excerpted from one transcript. The human-assigned importance of these importance score are 0.60, 0.40, and 0.70. For \emph{noise} and \emph{plan}, aggregated scores generated from word2vec-based embedding are 0.17 and 0.18, which does not belong to the same importance categories annotated. On the contrary, Bert-based embedding generates a score that aligns with human-assigned importance for \emph{noise} and \emph{plan}. However, for \emph{sunday}, the word2vec-based semantic score is relatively closer to the actual importance score than BERT-based embedding. In fact, \emph{sunday} appears as an isolated response to someone's question in transcript.}
\label{tab:model_parameters_9}
\end{table}

\subsection{Correlation Analysis to Assess Fit with Word Importance  Scores}

After calculating two types of embeddings for each word in this dataset, we asked which one would be more useful within a model to predict word importance. Prior work on the state-of-art word-importance learning algorithm Neural Bag-of-Words (NBOW) has revealed that learning importance of words within a sentence is effective while using the mean of each word-embedding vector as a feature \cite{sheikh:hal-01331720}. Following this common practice for determining word importance  \cite{kalchbrenner-etal-2014-convolutional, dai2020importance}, we calculated the mean of each word-embedding vector, to represent its word semantic score \cite{sheikh:hal-01331720}. For both the word2vec and BERT-based embeddings, for each sentence in the transcript, we normalized word-semantic scores within the sentence, to obtain a value in a [0,1] range for each word. BERT embeddings produce sub-word tokens for a complete word and to handle such a scenario we have computed the average of the sub-words to calculate the final composite semantic score.

After performing this operation across sentences in the transcripts, we conducted an analysis to determine which form of pre-trained embedding (word2vec or BERT) better correlated with human-produced annotations of word importance in the original dataset. The values based on word2vec were correlated with human annotations with a Pearson correlation coefficient of $r=0.30$, and for the BERT-based scores, the coefficient was $r=0.41$. A Fisher $z$-transformation \cite{upton2014dictionary} revealed that word semantic scores generated using BERT contextualized word embeddings were significantly better correlated ($z = -3.05, p<0.001$) with human-assigned scores than word2vec counterparts. Based on these findings, we decided to use BERT contextualized embeddings in continued analysis.

We also tried another traditional approach called TF-IDF to calculate a semantic score for words. A correlation analysis between the score generated by TF-IDF and human annotations resulted in a Pearson correlation coefficient of $r=0.25$, which was lower than the coefficient generated using word2vec word embedding.  

\section{Predicting Word Importance \vspace{-0.4em}}
To demonstrate how to use our dataset to predict the importance of each word, we have begun to investigate several supervised learning methods. The independent variable is the processed $ 768 \times 1$ BERT-embedding vector of each word, and the output variable is the human-labeled importance score, discretized into six classes, for each word in the dataset.  
\begin{table}[t]
\small
\centering

\begin{tabular}{|l|l|l|} 
\hline
    Method &  F1 Score & RMSE \\
\hline
    \begin{tabular}[l]{@{}l@{}}Multi-layer Perceptron\end{tabular}  & $0.10$ & $1.29$ \\
\hline
    \begin{tabular}[l]{@{}l@{}}Random-Forest\end{tabular}  & $0.25$ & $1.02$ \\
\hline
    \begin{tabular}[l]{@{}l@{}}Linear Support Vector \end{tabular}
    & $0.51$ & $0.99$ \\
\hline
    \textbf{Logistic Regression} & \textbf{$0.57$} & $0.92$ \\
\hline
 
\end{tabular}
\caption{  Supervised classification performance showing  macro-averaged F1 score and Root Mean Squared Error.}
\label{tab:model_parameters_2}
\end{table}
This classification experiment partitioned the corpus into 80\% training, 10\% development, and 10\% test set. This partition has been directly adapted from \cite{kafle-huenerfauth-2018-corpus}. We evaluated the model using two measures: (i) Root Mean Square Error (RMSE) - the deviation of the model predictions from the human-assigned categories, and (ii) the F1 measure for classification performance. For classification, we categorized annotation scores into the 6 levels, as described above: [0-0.1), [0.1-0.3), [0.3-0.5), [0.5-0.7), [0.7-0.9), and [0.9 - 1]. 

Table \ref{tab:model_parameters_2} illustrates that the better performing supervised model (of four traditional approaches) in predicting the importance class is Logistic Regression with F1-score $0.57$ and RMSE $0.92$. Even if the classes are discretized, we are generating continuous value for each word. And since both the human and supervised model generated scores, we calculated this RMSE. Among other approaches, the Linear Support Vector Classifier achieves F1-score 0.51, Random-Forest achieves 0.25, and Multi-layer Perceptron achieves 0.10.
\begin{table}[t]
\small
\begin{center}
\begin{tabular}{l|l|l|l|l|l|l|l|} 
     \multicolumn{8}{c}{Predicted Label}\\
    \cline{2-8} 
    & & 1 & 2 & 3 & 4 & 5 & 6  \\
  \cline{2-8} 
  \cline{2-8} 
  \multirow{5}{*}{\rotatebox{90}{True Label}}
  & 1 & \cellcolor[HTML]{BBBCCF} 0.69 & 0.21 & 0.18 & 0.15 & 0.18 & 0.00  \\
  \cline{2-8} 
  
  & 2 & 0.22 & \cellcolor[HTML]{CBBCCF}0.64 & 0.25 & 0.26 & 0.13 & 0.33  \\
  \cline{2-8} 
  & 3 & 0.05 & 0.12 & \cellcolor[HTML]{CCBCCF}0.48 & 0.11 & 0.18 & 0.00  \\
  \cline{2-8} 
  & 4 & 0.02 & 0.02 & 0.03 & \cellcolor[HTML]{CCBCCF}0.48 & 0.06 & 0.11  \\
  \cline{2-8} 
  & 5 & 0.01 & 0.01 & 0.04 & 0.00 & \cellcolor[HTML]{CCBCCF}0.40 & 0.00  \\
  \cline{2-8} 
  & 6 & 0.00 & 0.00 & 0.00 & 0.00 & 0.00 & \cellcolor[HTML]{CBBCCF}0.56 \\
  \cline{2-8} 
\end{tabular}
\caption{Normalized confusion matrices for Logistic Regression for classification into six word importance classes using BERT-generated embeddings-based score.}
\label{tab:model_parameters_3}
\end{center}
\end{table}

\section{Limitations and Future Work}
There are several limitations of this ongoing research that we intend to address in future work.
\begin{itemize}
    \item In our current research, we have determined a semantic score for each word using three methods. Future research can use other methods to generate the semantic score and retrospectively compare the generated semantic score with the score assigned by the human annotators.  
    
    \item The findings from this analysis leaves the room for future improvements, since we did not modify the hyperparameters to observe how accurately the models would predict the importance of words. Therefore, future research can explore variations of these models.
    
    \item Future directions may include collecting additional data to balance the distribution of importance classes. In addition, given the role of part of speech (POS) for word importance in texts \cite{Shah02astudy}, a next step could be to investigate POS with contextual word embedding for predicting word importance. Since TV captions often represent conversational speech with filler words, e.g., \emph{hmm} or \emph{yeah}, future research could consider alternative strategies to score the importance of such words. 
    
    \item \citet{hutchinson-etal-2020-social} and \citet{hassan-etal-2021-unpacking-interdependent} demonstrate that a large language model like BERT can introduce bias relating to people with disabilities into a task. Therefore, future work can investigate whether BERT is introducing any latent bias in predicting importance of words from DHH viewers' perspective.

\end{itemize}

% \section{Conclusion and Future Work}
\section{Conclusion}
The analysis presented above has revealed that BERT contextualized word-embedding can better represent the importance of words compared to word2vec embeddings, which had been used in prior work on word-importance prediction \cite{Kafle_ACE}. Research indicates that DHH viewers often follow key terms while skimming through captions, and researchers have proposed approaches to guide DHH readers to quickly identify keywords in caption text through visual highlighting  \cite{highlight_kafle_2019}. Our findings may allow broadcasters to use embeddings to determine the important words within a sentence and to highlight those words in captions, to support DHH viewers' ability to read \cite{amin-preference-2020} the captions effectively. In this study, a traditional Logistic Regression algorithm performed better at predicting importance classes.

We are also broadly investigating how to accurately measure the quality of caption transcriptions that are broadcast during live TV programs from the perspective of DHH viewers. We plan to incorporate predictive models into new word-importance weighted metrics, to better capture the usability of live captioning from DHH users' perspective.
\cite{amin-speakerid-2022}

\section{Ethics Statement}
This work advocates for improved inclusion of DHH individuals. A risk of the study is that results may not generalize across conversational corpora. 

\section*{Acknowledgments}
This material is based on work supported by the Department of Health and Human Services under Award No.\ 90DPCP0002-0100, and by the National Science Foundation under Award No. DGE-2125362. Any opinions, findings, and conclusions or recommendations expressed in this material are those of the author(s) and do not necessarily reflect the views of the Department of Health and Human Services or National Science Foundation.

\bibliography{acl}
\bibliographystyle{acl_natbib}

\end{document}